\pdfoutput=1

\documentclass[11pt]{article}

\usepackage{ACL2023}

\usepackage{times}
\usepackage{latexsym}

\usepackage[T1]{fontenc}

\usepackage[utf8]{inputenc}

\usepackage{microtype}

\usepackage{inconsolata}
\usepackage{graphicx}
\usepackage{float}

\usepackage{amsmath}
\usepackage{amssymb}
\usepackage{hyperref}
\usepackage{caption}
\usepackage{subcaption}

\usepackage{natbib}

\title{Is Anisotropy Inherent to Transformers?}

\author{
    Nathan Godey$\mkern6mu^{1,2}$ \quad Éric de la Clergerie$^1$ \quad Benoît Sagot$^1$ \\
    $^1$Inria, Paris, France \\
    $^2$Sorbonne Université, Paris, France \\
    \texttt{\{nathan.godey,eric.de\_la\_clergerie,benoit.sagot\}@inria.fr}
}

\begin{document}
\maketitle
\begin{abstract}
The representation degeneration problem is a phenomenon that is widely observed among self-supervised learning methods based on Transformers. In NLP, it takes the form of \textit{anisotropy}, a singular property of hidden representations which makes them unexpectedly close to each other in terms of angular distance (cosine-similarity). Some recent works tend to show that anisotropy is a consequence of optimizing the cross-entropy loss on long-tailed distributions of tokens. We show in this paper that anisotropy can also be observed empirically in language models with specific objectives that should not suffer directly from the same consequences. We also show that the anisotropy problem extends to Transformers trained on other modalities. Our observations tend to demonstrate that anisotropy might actually be inherent to Transformers-based models.
\end{abstract}

\section{Introduction}
In recent years, deep learning models based on Transformers have led to significant breakthroughs in the field of natural language processing (NLP). These models have demonstrated state-of-the-art performance across a range of tasks, such as language modeling, machine translation, and sentiment analysis. However, despite their successes, these models suffer from a phenomenon known as the representation degeneration problem. Specifically, this problem manifests as anisotropy, a property of hidden representations that makes them all close to each other in terms of angular distance (cosine-similarity).

Anisotropy has been widely observed among self-supervised learning methods based on Transformers, and recent research suggests that it may be a consequence of optimizing the cross-entropy loss on long-tailed distributions of tokens \citep{bis-etal-2021-much}. However, it remains uncertain whether anisotropy is a fundamental property of Transformers-based models or a consequence of the pre-training process.

In this paper, we investigate the anisotropy problem in depth, and we make several contributions:
\begin{itemize}
    \item We demonstrate empirically that anisotropy can be observed in language models with character-aware architectures that should not suffer directly from the same consequences as token-based NLP;
    \item We extend our observations to Transformers trained on other modalities, such as image and audio data;
    \item We provide empirical observations on the anisotropy problem in the Transformer block and argue that it may be an inherent property of the self-attention mechanism.
\end{itemize} 

\section{Related Work}

\begin{figure}[h]
     \includegraphics[width=\columnwidth]{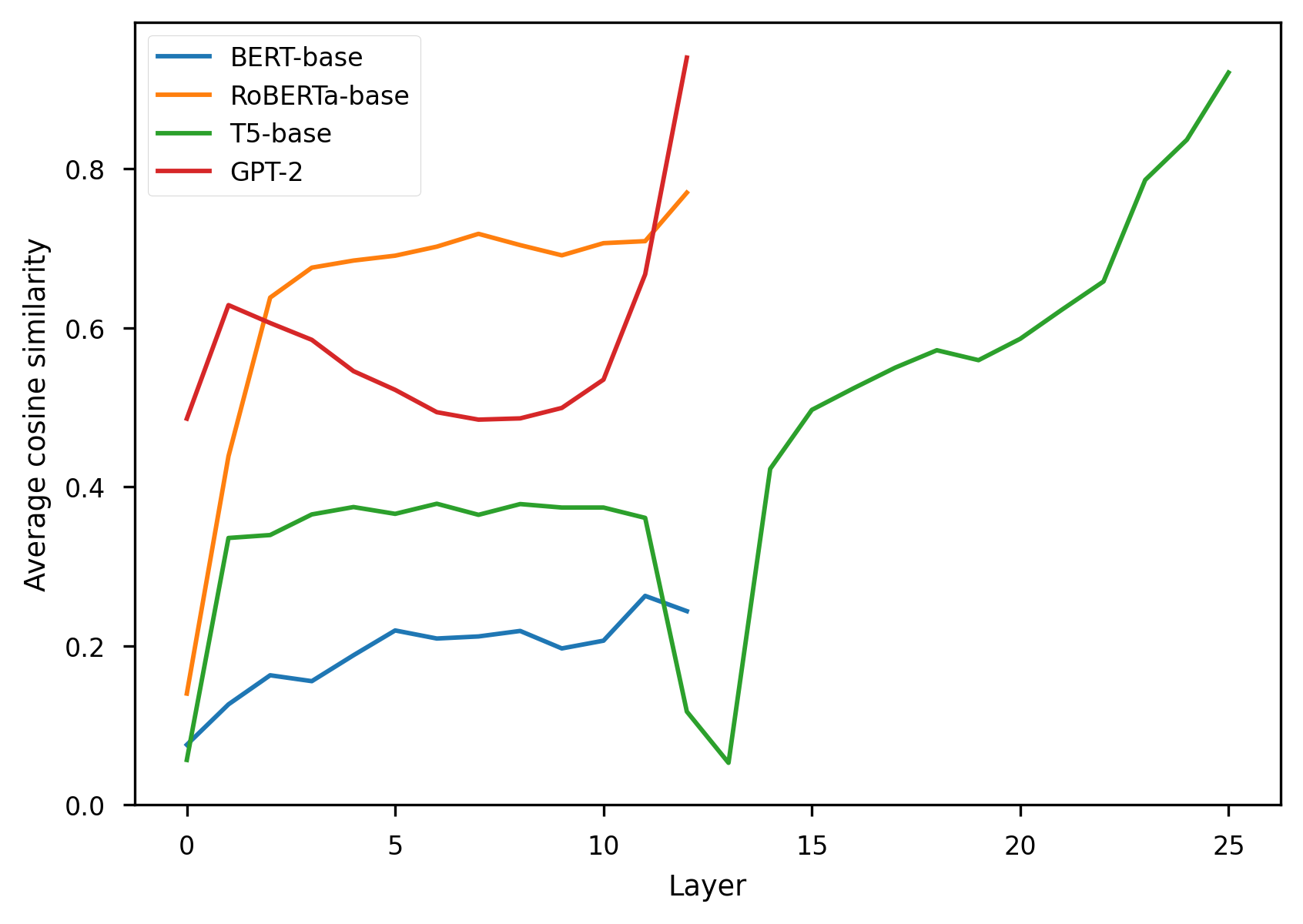}
     \caption{Average cosine-similarity between hidden representations across layers for token-level NLP models. For T5-base, we concatenate encoder and decoder results.}
     \label{fig:anisotropy_token}
\end{figure}

The general phenomenon of anisotropy in token-based Transformers for language models has been shown in \citet{ethayarajh-2019-contextual}. \autoref{fig:anisotropy_token} extends one of their experiment to more architectures. \citet{GaoHTQWL19} shows that the degeneration of representations come from the distributions of subwords in natural language, namely the existence of unused and rare tokens, that tend to push all representations away from the origin towards a specific direction.

Other works have established a connection between word frequency and distortions of the latent spaces \citep{yu-etal-2022-rare, puccetti-etal-2022-outlier, rajaee-pilehvar-2022-isotropy}. \citet{bis-etal-2021-much} have shown that anisotropy in LMs could be explained by a global \textit{drift} of the representations in the same direction, thus unifying conclusions from \citet{ethayarajh-2019-contextual} and \citet{GaoHTQWL19}. The authors propose that this drift is caused by the persistent updating of the representation of rare and unused tokens in a consistent direction, due to the nature of the softmax operation in the cross-entropy loss. They show that removing the average component to all representations leads to a nearly perfect isotropy.

Several methods have been proposed to reduce anisotropy in Transformers-based LMs at token-level \citep{rajaee-pilehvar-2021-cluster, Wang2020Improving}, or at sentence-level \citep{gao-etal-2021-simcse, yan-etal-2021-consert, su2021whitening}. They usually consist in post-processing the representations, and lead to downstream performance boosts. We argue that these positive results are paving the way for the search of pre-training objectives that do not introduce anisotropy in the first place, in the hope that the resulting models will also be more performant without any post-processing, and potentially be trained more efficiently.

\section{Anisotropy in pre-trained Transformers}
\subsection{Character-based NLP}
\label{sec:charbased}
\begin{figure}[h]
     \includegraphics[width=\columnwidth]{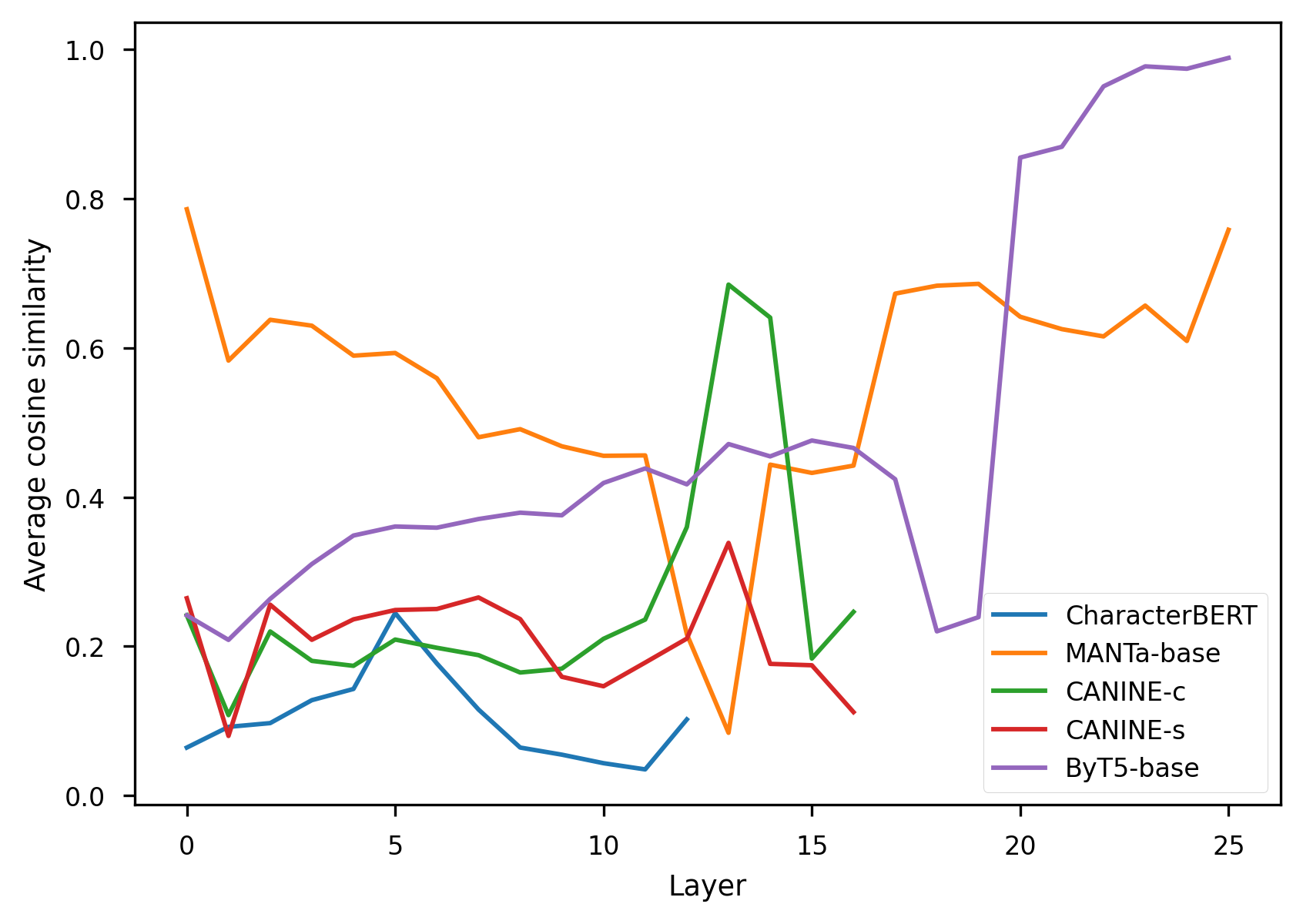}
     \caption{Average cosine-similarity between hidden representations across layers for character-level models.}
     \label{fig:cos_char_aware}
\end{figure}

To validate that the cross-entropy objective applied on vocabularies containing rare tokens is the main cause for the common drift issue, we explore anisotropy in character-based models. We study different architectures:
\begin{itemize}
    \item CharacterBERT \citep{el-boukkouri-etal-2020-characterbert} is constructing whole word representations from character embeddings put through convolutions and highway layers, before feeding them to a Transformers architecture.
    \item CANINE \citep{clark-etal-2022-canine} is downsampling contextualized character representations via a strided convolution before feeding them to a Transformers. It can be trained either with a subword-based objective (CANINE-s) or with a character-level one (CANINE-c).
    \item MANTa-LM \citep{godey-etal-2022-manta} is based on a differentiable segmentation and embedding module added before an encoder-decoder model in the style of T5 \citep{2020t5}. It takes bytes as inputs and outputs, but builds internal representations that are usually based on several bytes.
    \item ByT5 \citep{xue-etal-2022-byt5} is a version of T5 that is trained at byte-level. To afford for more complex encoding, the authors resize the encoder-decoder architecture.
\end{itemize}

Neither of these architectures should suffer from out-of-vocabulary in the process of creating representations. The models that predict at word or sub-word level (CharacterBERT and CANINE-s) could have the cross-entropy loss systematically pushing away rare item representations. However, it is rather unclear why it would imply an embedding drift at deeper layers. Hence, if anisotropy was only caused by the presence of unused or rare subwords, those character-level models should be much less prone to this issue.

To verify this hypothesis, we compute hidden representations for the validation set of the WikiText-103 corpus \citep{MerityXBS16}. We then compute the average cosine-similarity between two representations, uniformly taken in the whole validation corpus.

In fact, as we can see in \autoref{fig:cos_char_aware}, those models all display significant levels of anisotropy at least in one of their layers. Interestingly, the models that are based solely on characters or bytes for input and prediction (ByT5 and CANINE-c) seem to display even higher levels of anisotropy. We note, as it is the case for the T5 model, that the ByT5 decoder displays extremely high levels of anisotropy.

\subsection{Other modalities}
\label{sec:other_mod}
\begin{figure*}[h]
    \centering
    \begin{subfigure}[b]{0.48\textwidth}
         \includegraphics[width=\linewidth]{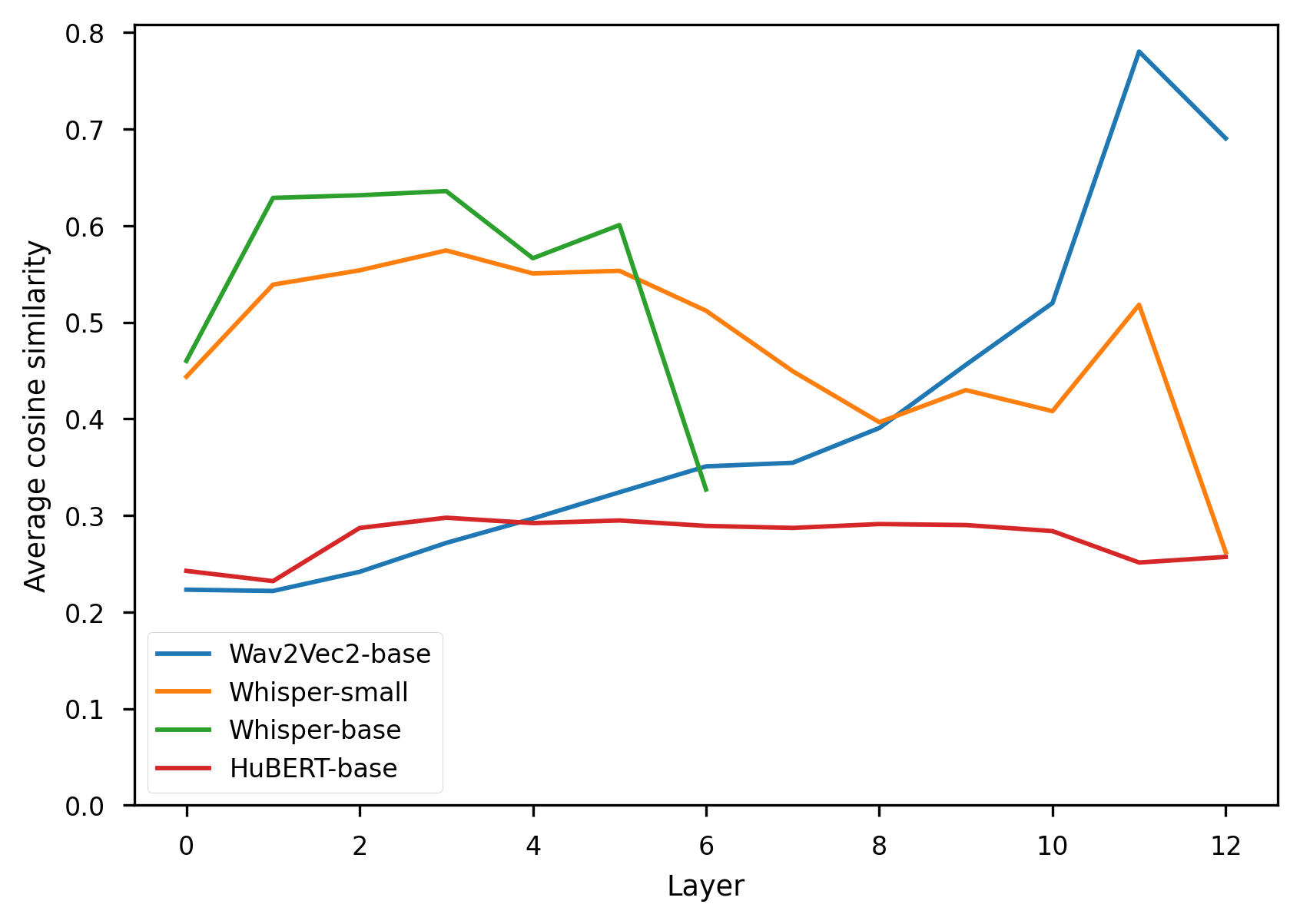}
         \caption{Speech}
         \label{fig:cos_speech}
    \end{subfigure}
    \begin{subfigure}[b]{0.48\textwidth}
         \includegraphics[width=\linewidth]{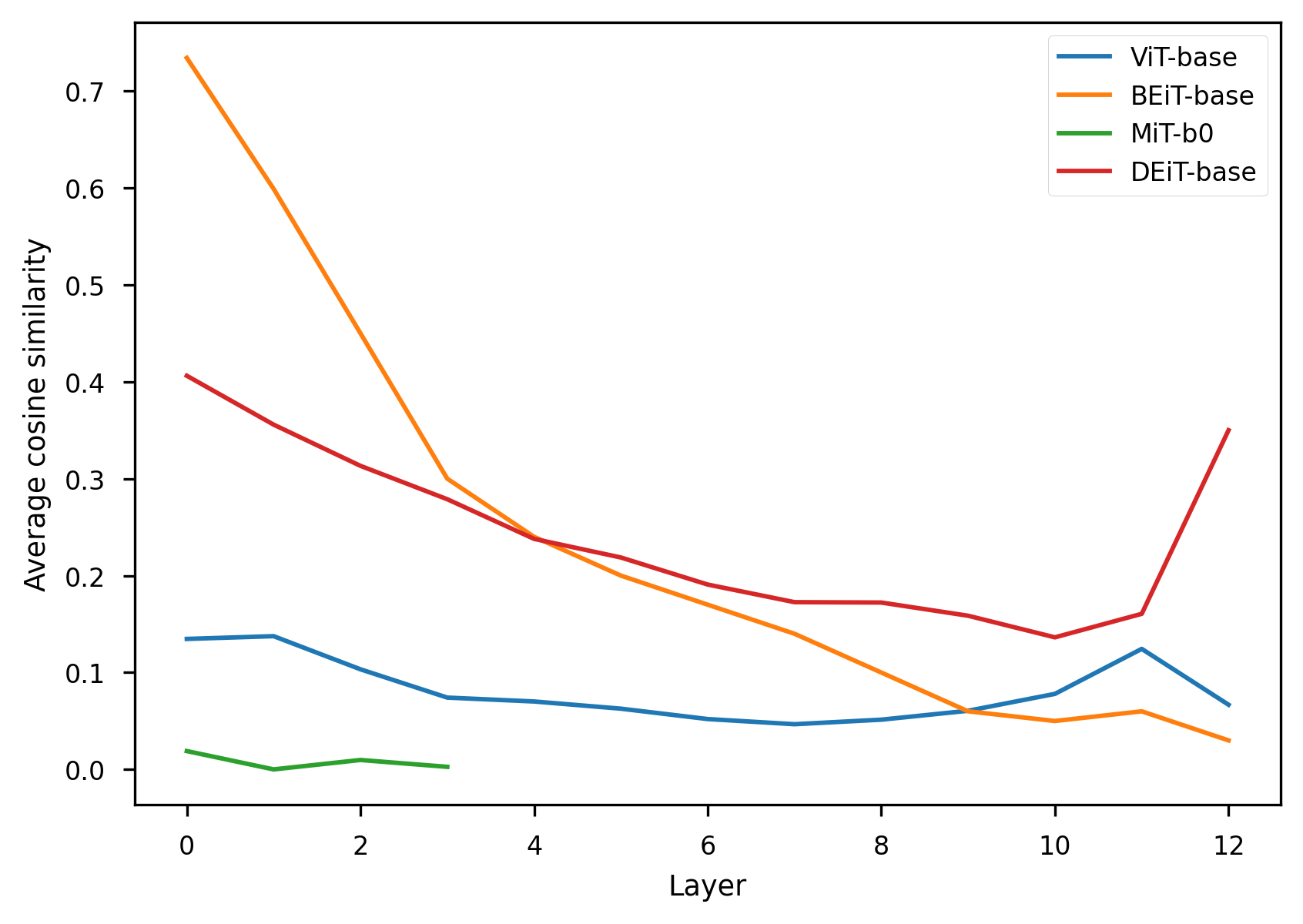}
         \caption{Vision}
         \label{fig:cos_audio}
    \end{subfigure}
    \caption{Average cosine-similarity between hidden representations across layers for Speech and Vision modalities. We observe that across both modalities, several models display significant levels of anisotropy.}
    \label{fig:anisotropy_modalities}
\end{figure*}
We've shown in the previous section that character-level language models suffer from anisotropy similarly to token-level ones. We proceed to explore the anisotropy problem for Transformers-based models in other modalities, specifically speech and vision.

For speech models, we consider wav2Vec 2.0 \citep{wav2vec}, HuBERT \citep{HuBERT}, and Whisper \citep{radford2022whisper} with the Common Voice 11.0 dataset \citep{commonvoice:2020}. For vision models, we use ViT \citep{Wu2020VisualTT}, BEiT \citep{beit-2021}, MiT \citep{segformer21}, and DEiT \citep{pmlr-v139-touvron21a} on the ImageNet dataset \citep{imagenet15russakovsky}.

As in \autoref{sec:charbased}, we infer hidden representations on the validation sets for each modality. We then uniformly sample pairs of vectors to get cosine-similarity values for every layer of every model. The results are displayed in \autoref{fig:anisotropy_modalities}.

Once again, almost every model shows a significant level of anisotropy on some of its layers. Notably, speech models show very anisotropic representations, as every layer of every model outputs an average cosine-similarity of at least $0.2$. We find some exceptions among vision models, since the MiT model seems to use only isotropic representations and the ViT model has a low average cosine-similarity for all its layers.

We also conduct the same experiment for convolution-based networks in the vision modality. The models at glance are ResNet \citep{he2016deep}, EfficientNet \citep{Tan2019EfficientNetRM}, CvT \citep{wu2021cvt}, ConvNeXt \citep{liu2022convnet}, and VAN \citep{guo2022visual}. For these networks, we flatten convolution maps before computing the cosine-similarity.

\begin{figure}[H]
    \centering
    \includegraphics[width=\linewidth]{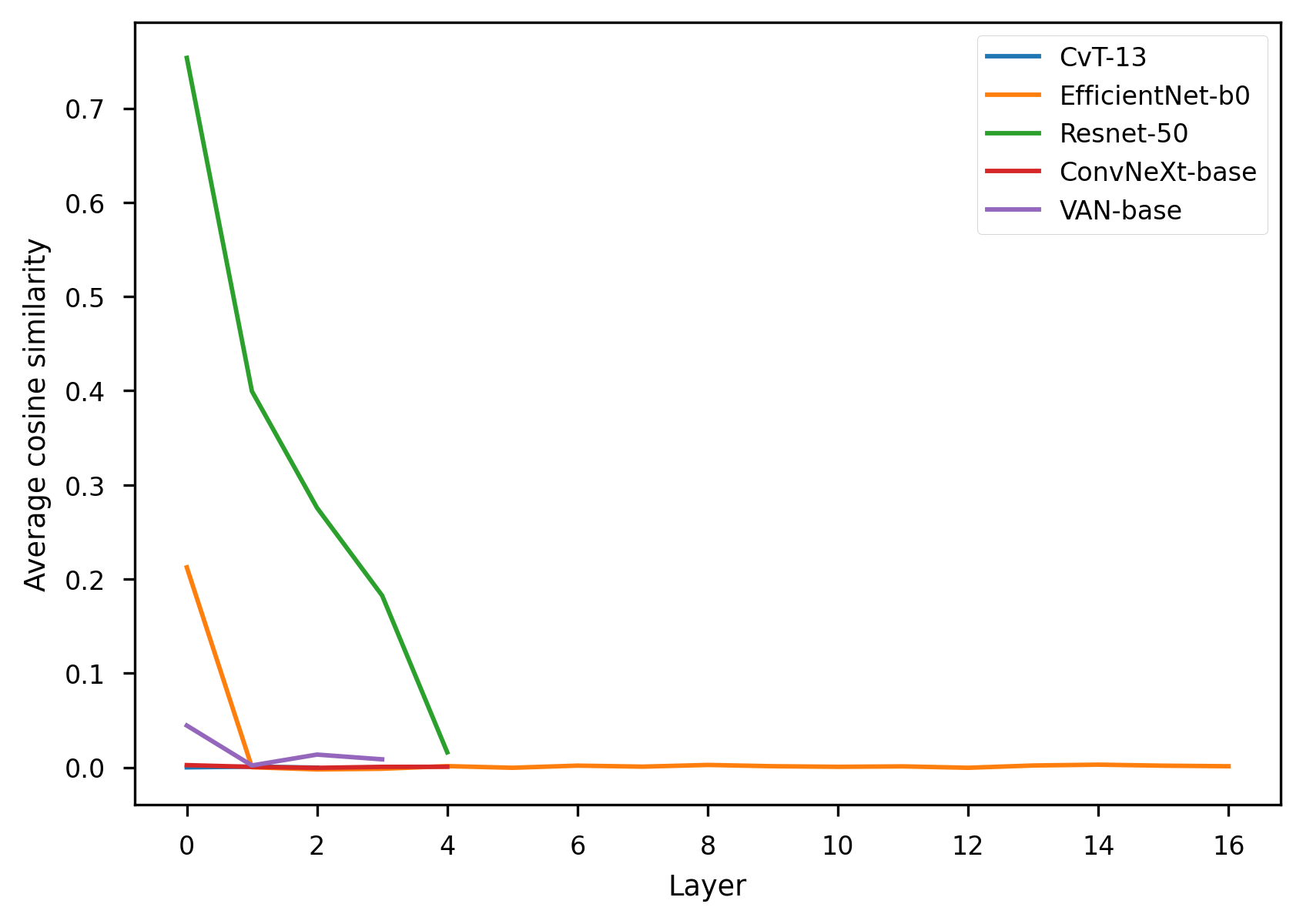}
    \caption{Average cosine-similarity between hidden representations across layers for convolution-based vision models.}
    \label{fig:convbased}
\end{figure}

We observe in \autoref{fig:convbased} that most of the convolution-based models are isotropic. Interestingly, the only exception is ResNet-50, whose representations become more and more isotropic as one explores deeper layers. This could partially be explained by the fact that the batch normalization \citep{pmlr-v37-ioffe15} used in some of these models mitigates \textit{a posteriori} the drift effect by removing the mean component of the representations. However, the ConvNeXt model also seems to use isotropic representations while not using batch normalization, which shows that this is not the only factor in the isotropic behavior of these models.

\subsection{To drift or not to drift?}
Related works \citep{bis-etal-2021-much, GaoHTQWL19} show that anisotropy in subword-level language models is caused by a drift of the hidden representations in a shared direction. In this section, we try to extend this observation to other modalities.

We study the correlation between the uniformly measured cosine-similarity, and the norm of the average hidden representation $||\bar{x}||_2$ for each layer. If anisotropy could be explained by the drift effect, we would expect a monotonic relation between $||\bar{x}||_2$ and the average cosine-similarity. To verify this, we apply a Spearman correlation test on these two metrics for every model from \autoref{sec:charbased} and \autoref{sec:other_mod}, along with some token-level language models, namely T5 \citep{2020t5}, BERT \citep{devlin-etal-2019-bert}, RoBERTa \citep{roberta}, and GPT-2 \citep{gpt2}.

\begin{figure}[h]
    \centering
    \includegraphics[width=\linewidth]{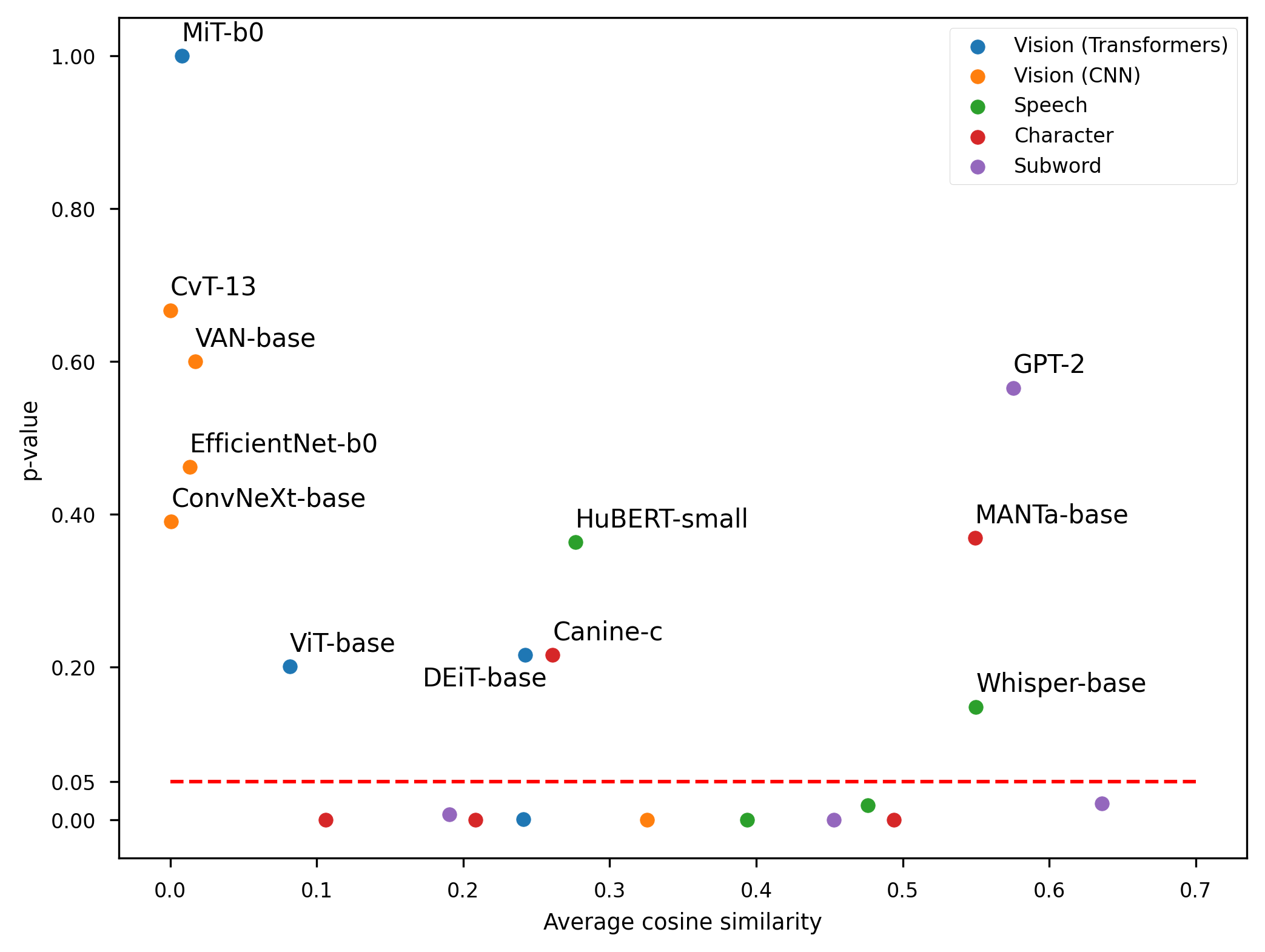}
    \caption{p-value of the Spearman correlation test between the norm of the average representation and the cosine-similarity averaged over all layers, across modalities. Models above the red dotted line are not significantly affected by the drift effect.}
    \label{fig:pval_vs_cos}
\end{figure}

In \autoref{fig:pval_vs_cos}, we observe that we cannot correlate the anisotropy level and the drift component for several models. We notice that the anisotropy affecting some CNN-based vision models can be explained by the drift effect, but we underline that this is not the case for Tranformers-based models in the same modality. Some speech models (HuBERT and Whisper-base) also display signs of anisotropy that cannot be correlated with the drift effect. On the other hand, the anisotropy of subword-based models can generally be explained by the drift effect, except for GPT-2 for which the Spearman correlation metric may not be appropriate. We provide a similar analysis based on the Pearson correlation test and discuss the relevance of each statistic in \autoref{sec:pearson}.

Interestingly, \autoref{fig:pval_vs_cos} displays a correlation for all character-based models but Canine-C and MANTa-base. All in all, if \autoref{fig:pval_vs_cos} confirms the conclusions of related works when it comes to token-based language models, it contradicts the hypothesis that the anisotropy in Transformers-based models can be generally be explained by the drift effect.

\section{Explaining the representation drift}
\label{sec:empirical}
In this section, we expose some properties of the Transformer block as used in BERT \citep{devlin-etal-2019-bert}. We analyze experimentally the behavior of the Transformer block $T$ when a common bias term $b$ is added to the input representations $\mathbf{x}$. This allows us to mimic the common drift as mentioned in \citet{bis-etal-2021-much} and to identify some properties induced by this artificial drift on the output representations.

\subsection{Experimental setup}
We consider an embedding lookup table $E$ and a Transformer block $T$ built and initialized as in BERT \citep{devlin-etal-2019-bert}. We then draw 16 input embedding sequences $\mathbf{x}$ of length 512 uniformly from $E$. To account for a drift component of norm $N\in\mathbb{R}$, we generate a vector $b_u \sim \mathcal{N}(0, I_d)$, which we normalize into $b = \frac{b_u}{||b_u||_2}\times N$. We finally compute $T(\mathbf{x}_i + b)$ for every sequence $x_i$, and study the resulting distributions.

Specifically, we study the average norm of the input representations $E(||\mathbf{x}_i + b||_2)$ against the average norm of the output representations $E(||T(\mathbf{x}_i + b)||_2)$ in \autoref{fig:norm_scratch_transformer}. We also retrieve the self-attention scores before the softmax operation, namely $\frac{QK^T}{\sqrt{d_k}}$, along with the corresponding $Q$ and $K$ matrices. We study some of their properties in \autoref{fig:attscore_trained_transformer} and \autoref{fig:kq}.

\subsection{Input vs. output analysis}
\begin{figure}[h]
    \centering
    \begin{subfigure}[b]{\columnwidth}
         \includegraphics[width=\linewidth]{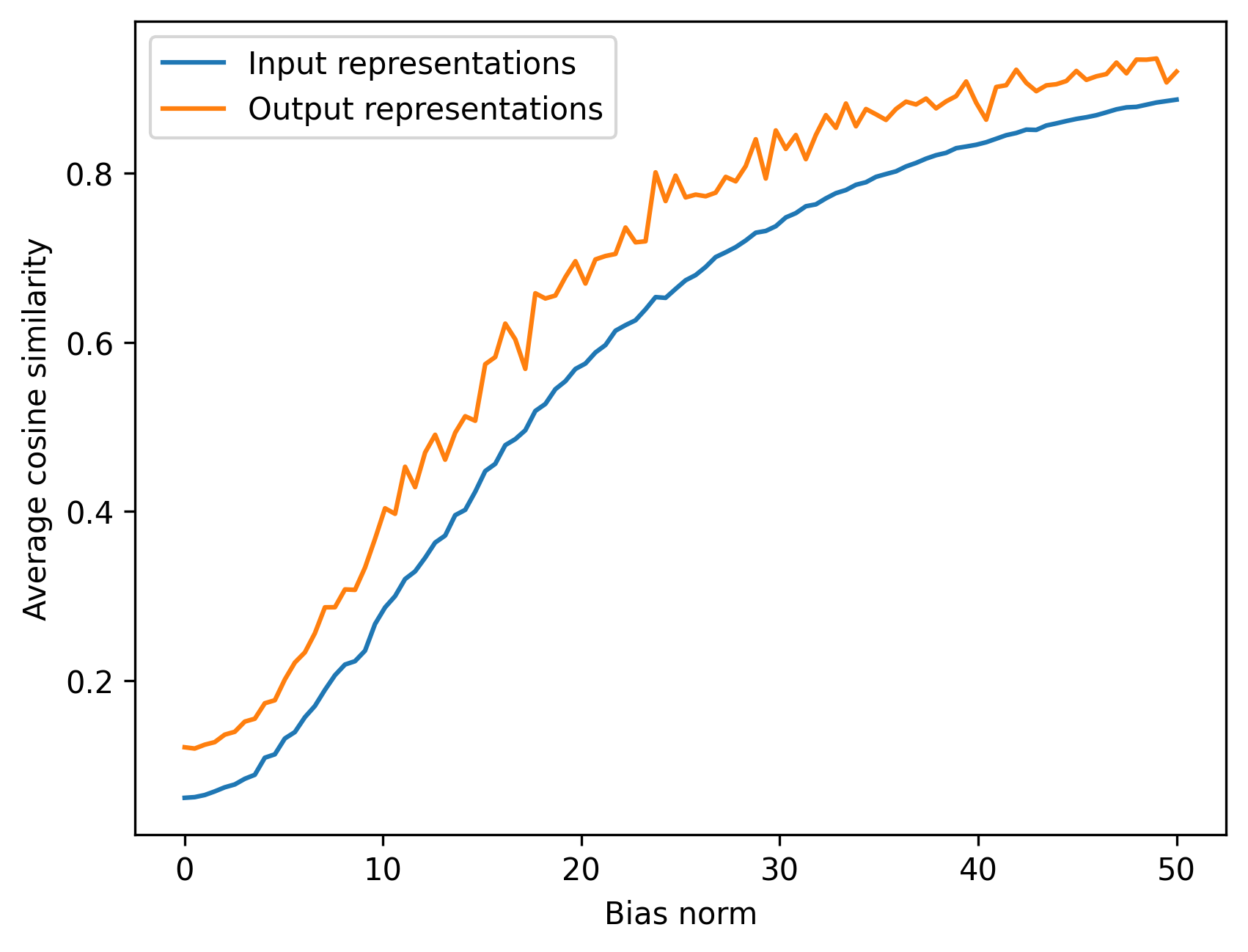}
         \caption{Cosine similarity}
         \label{fig:cos_scratch_transformer}
    \end{subfigure}
    \begin{subfigure}[b]{\columnwidth}
         \includegraphics[width=\linewidth]{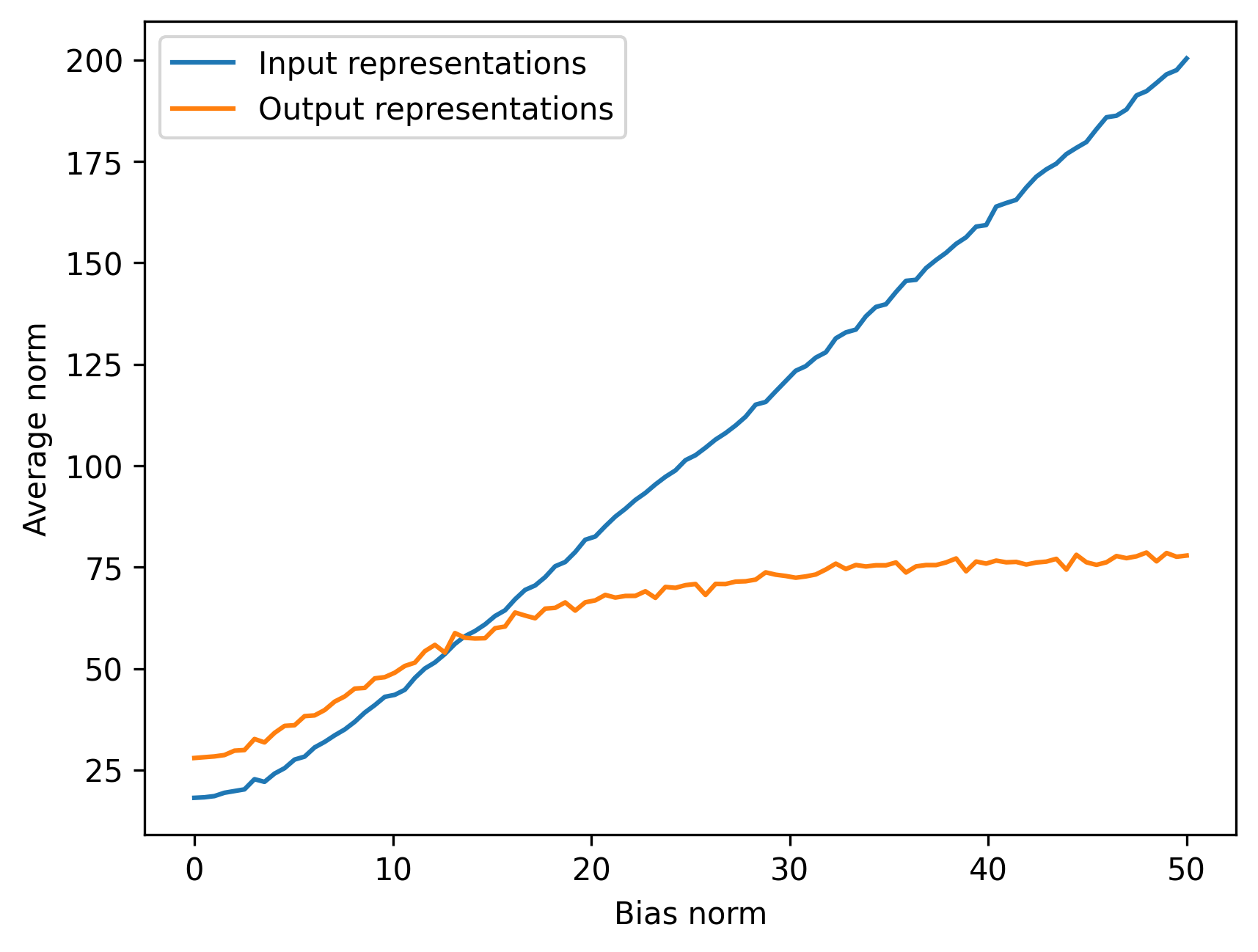}
         \caption{Norm}
         \label{fig:norm_scratch_transformer}
    \end{subfigure}
    \caption{Input/Output comparison of a Transformer block from BERT-base as the bias norms increases.}
    \label{fig:bias_vs_cosine_norm}
\end{figure}

In \autoref{fig:cos_scratch_transformer}, we observe that the output representations have an average cosine-similarity value that is slightly higher than the one of the input representations, no matter the level of input bias. We also notice that while the norm of the average output representation increases with the bias norm, it seems to meet the corresponding input measure for a given bias norm.

Interestingly, this shows that there is a fixed point in terms of norm in the Transformers function with biased input. More formally, there seems to exist a bias norm $N^* \in \mathbb{R}_+$ such that $$E_{x, b_{N^*}}(||x_i + b_{N^*}||) = E_{x, b_{N^*}}(||T(x_i + b_{N^*})||)$$

Moreover, this fixed point level $N^*$ is in the order of magnitude of the average hidden state norms of the layers of the trained BERT model. This hints that the model's representations stabilize when their norm is close to this fixed point.

\subsection{Exploring the Transformer block}

To understand the effect of the drift effect on the inner workings of the Transformer layer, we take a closer look at the self-attention operation as the average input representation drifts away.

\begin{figure}[h]
    \centering
    \includegraphics[width=\linewidth]{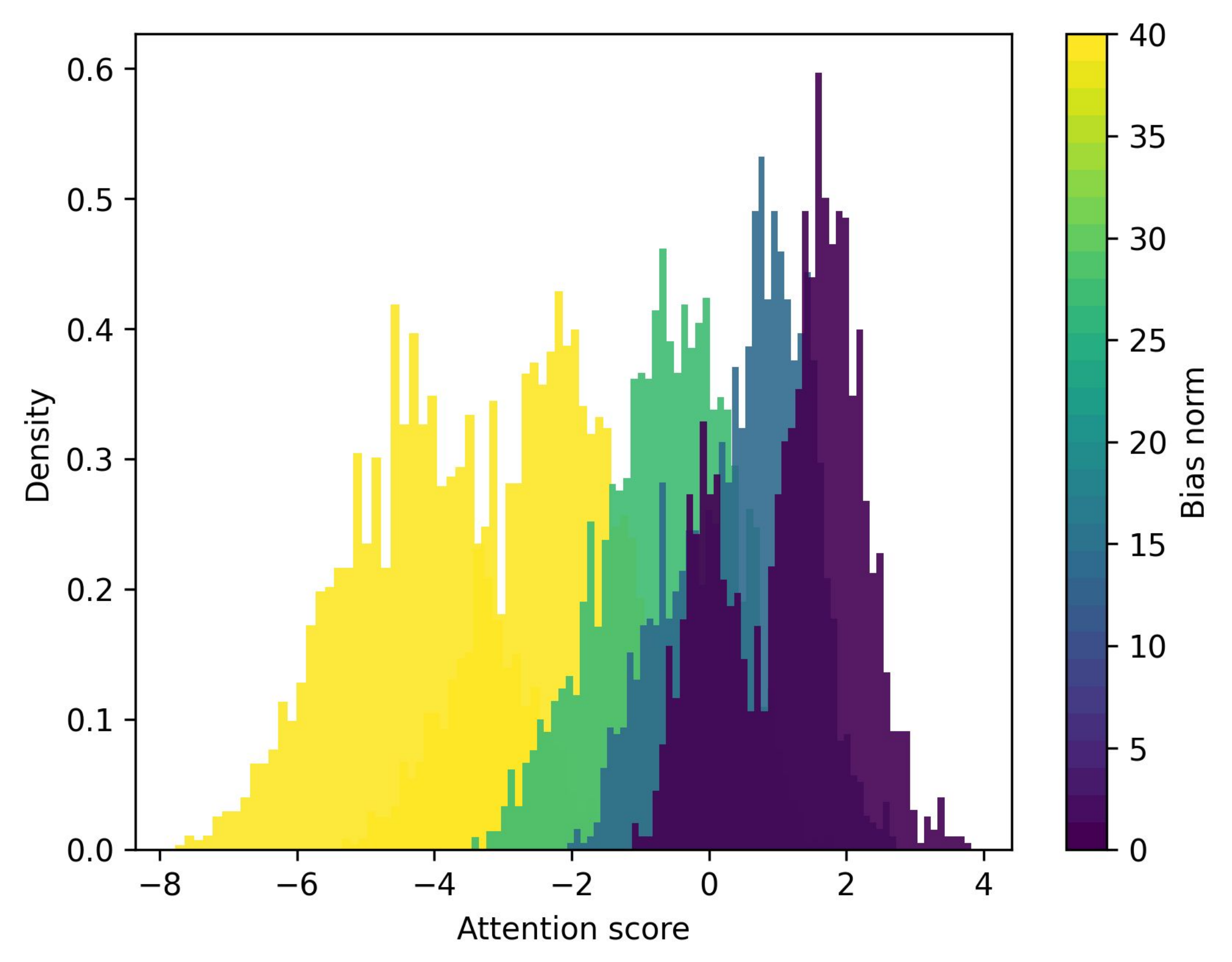}
    \caption{Histograms of the pre-softmax attention scores as the input bias norm increases. Other initializations of the layer and of the bias direction $b_u$ led to a general \textit{increase} of the attention scores instead.}
    \label{fig:attscore_trained_transformer}
\end{figure}

\autoref{fig:attscore_trained_transformer} shows that the attention scores tend to move away from zero as the input bias norm increases. Indeed, as the norm of the average $\bar{x}$ of the input embeddings increases, we can expect the query and key vectors $Q$ and $K$ to also display signs of anisotropy. Actually, for each self-attention head, and for all position $i \in [1, L]$, one should observe:
\begin{equation}
    \begin{cases}
      E_x(Q_i) = W_Q\bar{x} + b_Q\\
      E_x(K_i) = W_K\bar{x} + b_K
    \end{cases}
\end{equation}

\begin{figure}[h]
    \centering
    \begin{subfigure}[b]{0.48\columnwidth}
         \includegraphics[width=\linewidth]{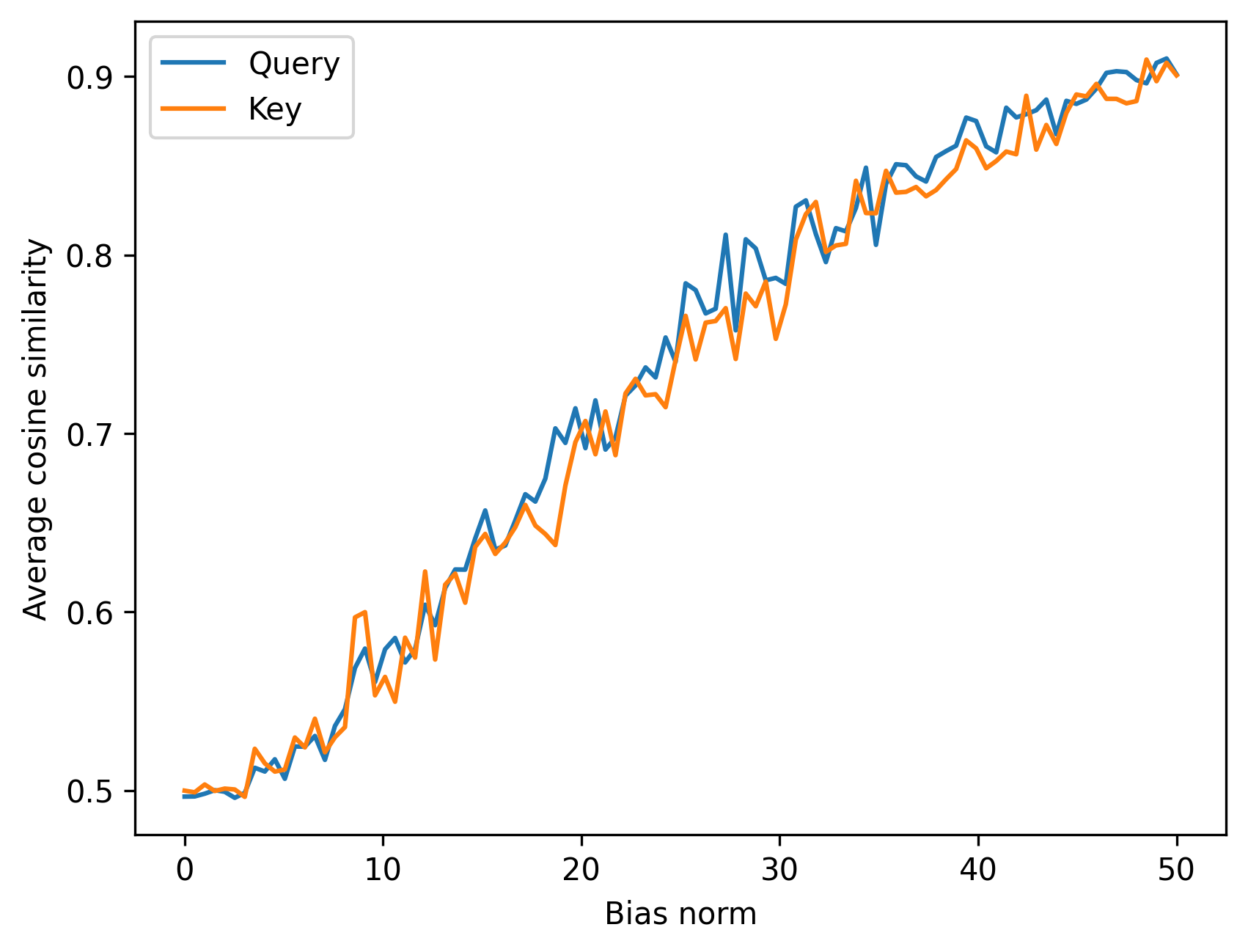}
         \caption{Cosine sim.}
         \label{fig:cos_qk_trained_transformer}
    \end{subfigure}
    \begin{subfigure}[b]{0.48\columnwidth}
         \includegraphics[width=\linewidth]{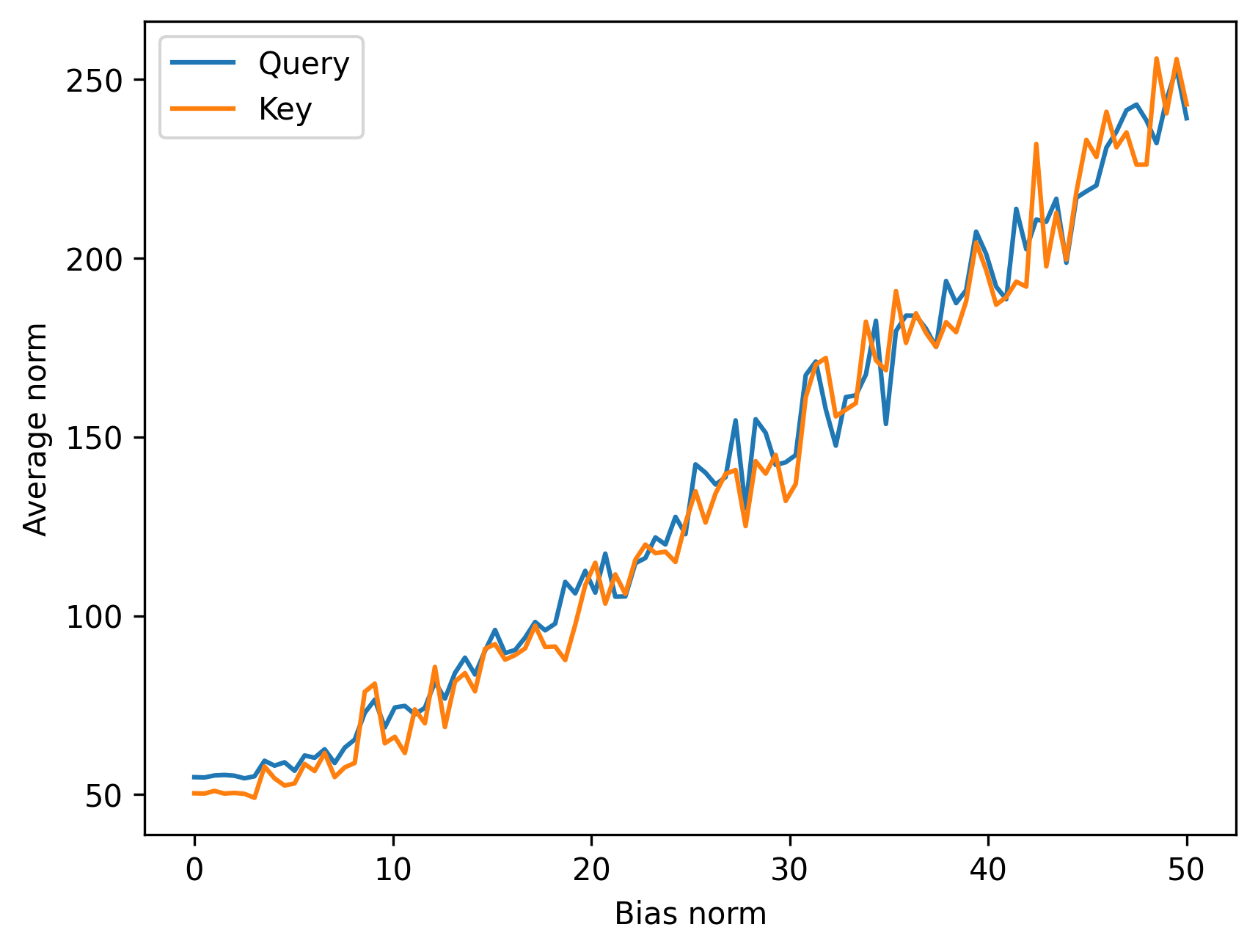}
         \caption{Norm}
         \label{fig:norm_qk_trained_transformer}
    \end{subfigure}
    \caption{Analysis of the self-attention query and key distributions}
    \label{fig:kq}
\end{figure}

We can observe in \autoref{fig:kq} that query and key representations indeed increase in norm with the input bias norm. We also notice that the corresponding distributions are anisotropic even when no bias is added, which may be a consequence of BERT's initialization parameters.

\subsection{Impact of the drift}

After exploring the consequences of the drift of input representations on the query-key product in self-attention, we identify in this section the implications of this drift at a more explainable level.

\begin{figure}[h]
    \centering
    \includegraphics[width=\linewidth]{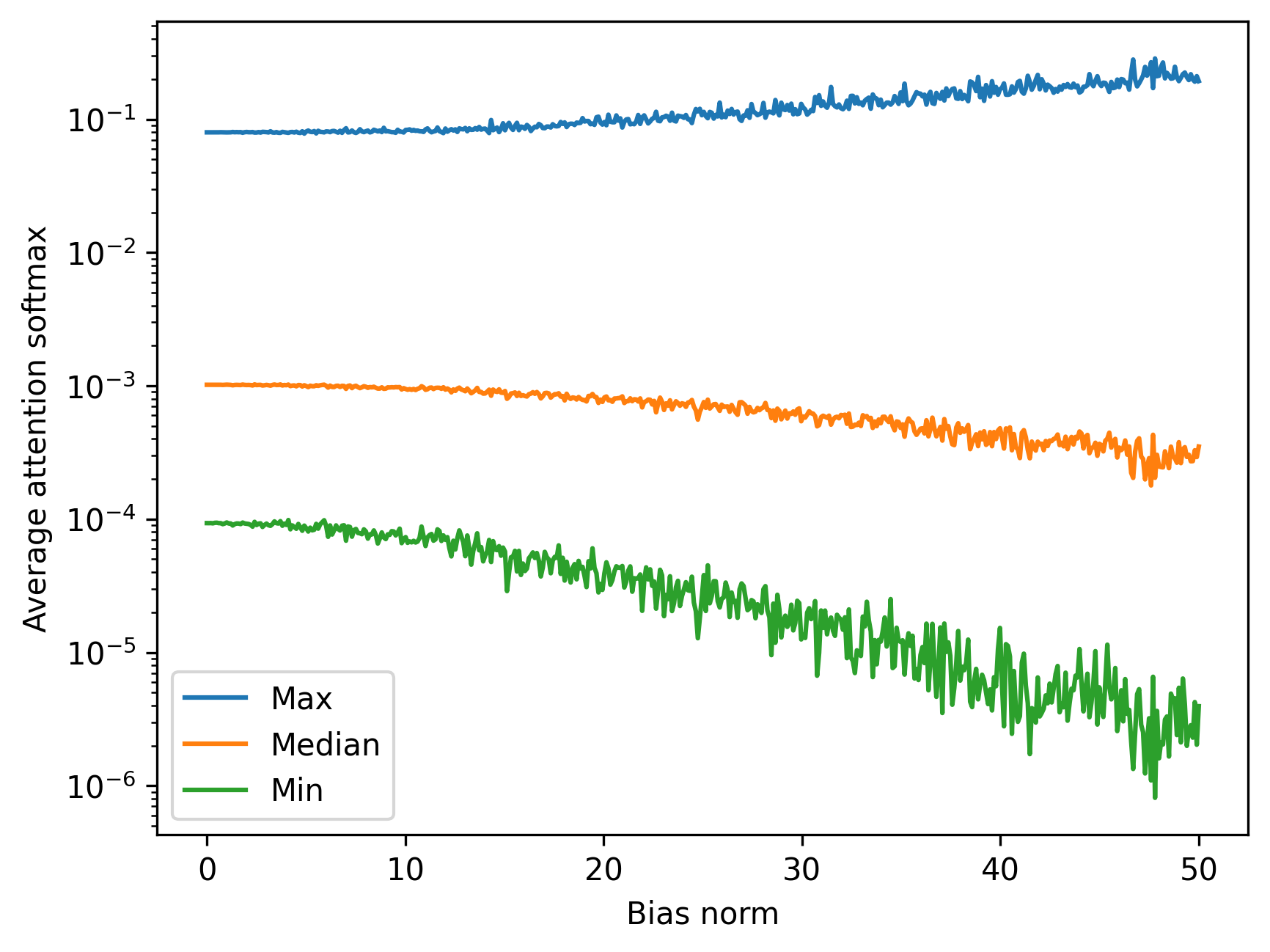}
    \caption{Evolution of the self-attention softmax values as the input bias norm increases.}
    \label{fig:softmax_trained_transformer}
\end{figure}

In \autoref{fig:softmax_trained_transformer}, we retrieve softmax values in the self-attention block and for each position, we extract the maximum, the median and the minimum. We then average these values over the whole batch, and repeat for various input bias norm levels. We notice that as the input bias norm increases, the self-attention softmax distributions become more and more categorical.

\begin{figure}[h]
    \centering
    \begin{subfigure}[b]{0.48\columnwidth}
         \includegraphics[width=\linewidth]{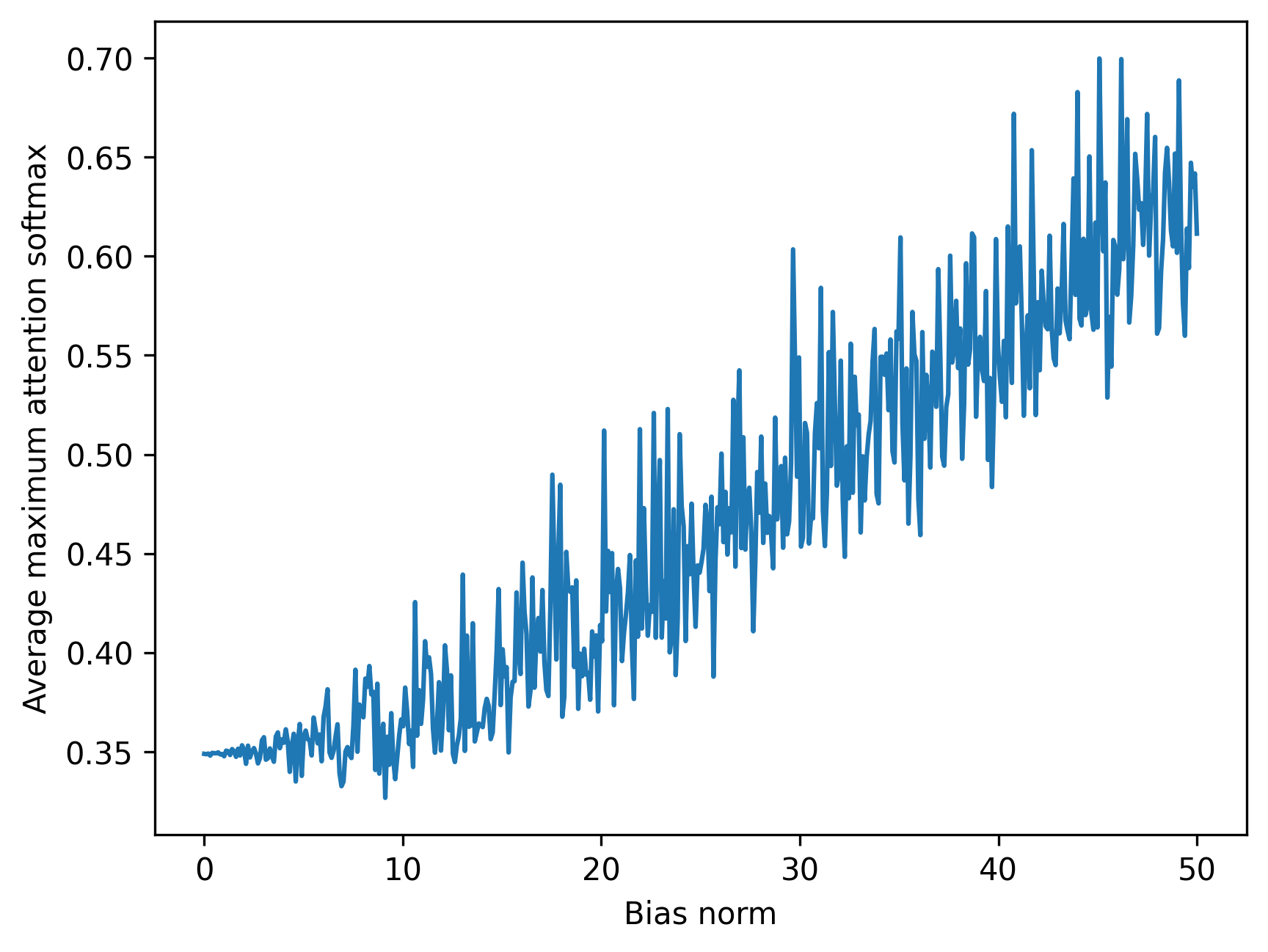}
         \caption{Maximum}
         \label{fig:max_softmax}
    \end{subfigure}
    \begin{subfigure}[b]{0.48\columnwidth}
         \includegraphics[width=\linewidth]{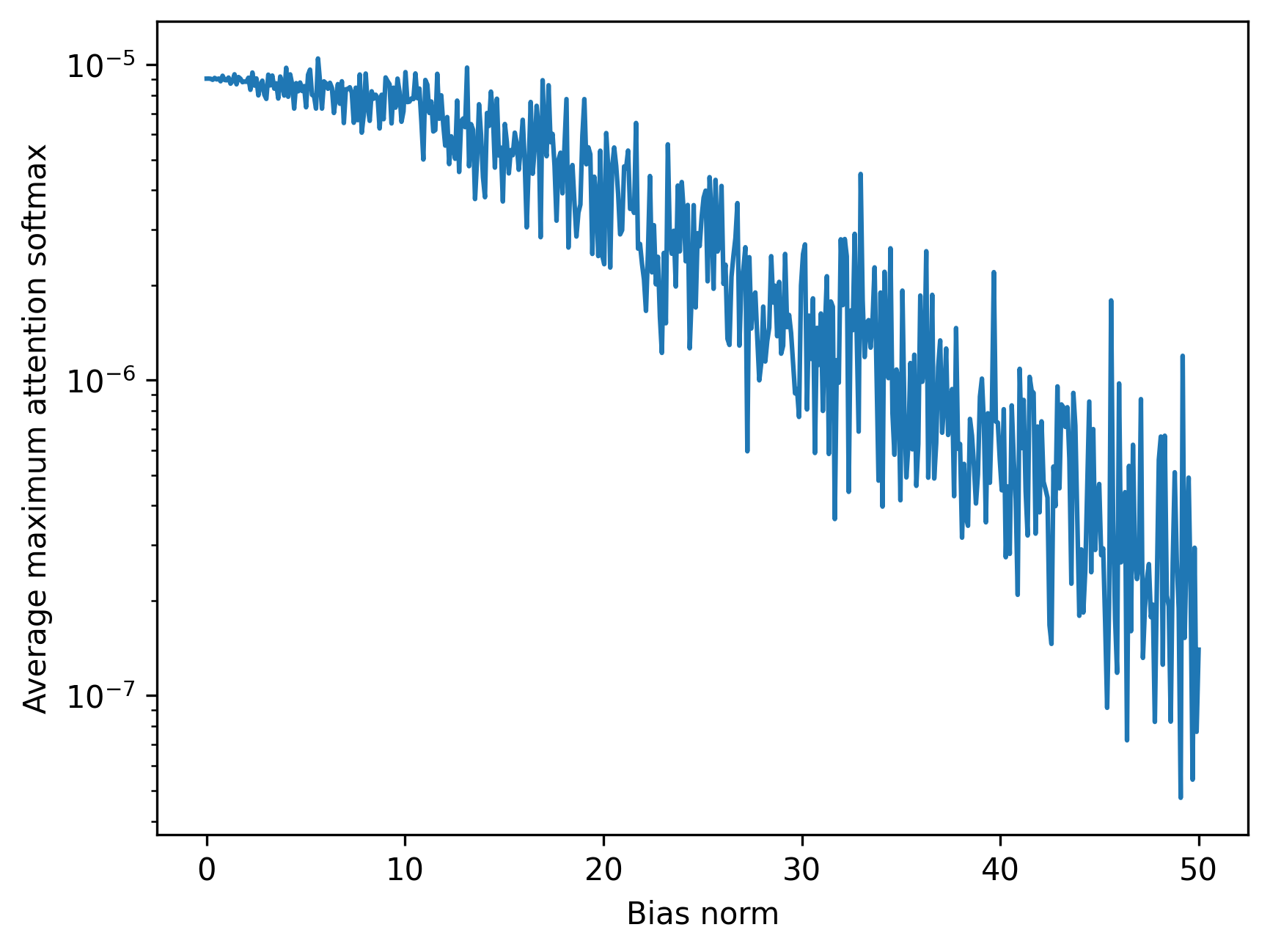}
         \caption{Minimum}
         \label{fig:min_softmax}
    \end{subfigure}
    \caption{Comparison of the extreme values of each sequence averaged over the batch as the bias norm increases.}
    \label{fig:min_vs_max}
\end{figure}

This becomes even clearer if we take the maximum and minimum values over the whole sequences, as in \autoref{fig:min_vs_max}.

However, at low anisotropy levels, i.e. when the bias norm is low, we see that the effect is not very important. \autoref{fig:softmax_trained_transformer} and \autoref{fig:min_vs_max} only hint at the fact that the drift of embeddings may help the self-attention to be more categorical. Another explanation could be that training favors categorical self-attention patterns, as has been pointed out in previous works \citep{clark-etal-2019-bert}, which in turn induces a drift in the models' representations. We leave these questions for future work.

\section{Discussion}
In this work, we hypothesize that the drift effect in Transformers-based models is not solely responsible for the anisotropy observed in most hidden representations across modalities. As \autoref{sec:empirical} shows, untrained Transformers layers display a tendency towards anisotropy. Biased inputs tend to increase the variance of the attention scores and thus facilitate the emergence of categorical patterns in the self-attention mechanisms.

To strengthen our hypothesis, an analysis of anisotropy via drift during the pre-training process would help understand how the query and key distributions evolve and verify that the progressive drift of the input representations leads to a progressive increase of the sharpness of the self-attention behavior. We found that such a study would require a more complex experimental setup, as the input data distribution would then need to be considered carefully, and as the Gaussian sample of a bias term becomes less relevant since a specific drift direction naturally emerges in the representations.

We hope that this work paves the way for a clearer understanding of the inherent anisotropy in Transformer-based models. Even though anisotropy has not been shown to be an issue in language modeling, previous works have advocated that removing anisotropy in output representations leads to better sense disambiguation abilities \citep{bihani-rayz-2021-low, bis-etal-2021-much}. It appears to us that re-thinking the self-attention operation to facilitate the emergence of categorical patterns with minimal implications for the input representations could be an interesting continuation of our work.

\section*{Conclusion}
In this paper, we investigated the anisotropy problem through the lens of the drift effect, and made several contributions to the understanding of this phenomenon. We demonstrated that anisotropy can be observed in language models with character-aware architectures, extended our observations to Transformers trained on other modalities, and studied anisotropy in untrained Transformers layers.

We conclude that anisotropy almost systematically affects Transformers on all modalities, in a way that is not always correlated with the drift of the representations. We also provide empirical evidence that anisotropy may be an inherent property of the self-attention mechanism when modeling categorical attention patterns. We hypothesize that a revision of the self-attention operation could help reduce anisotropy by facilitating the emergence of sharp self-attention softmax distributions without modifying the geometry of the hidden representations.

\section*{Limitations}
As mentioned in the Discussion section, we acknowledge that \autoref{sec:empirical} does not take into account the training dynamics, and only exposes some properties of the Transformer layer at initialization. We also notice that the Spearman correlation test used in \autoref{fig:pval_vs_cos} may not be well-suited for such noisy observations, as the high p-value of the GPT-2 model shows. We provide a similar graph based on the Pearson correlation in \autoref{sec:pearson}.

\section*{Ethics Statement}
To the best of our knowledge, our work does not raise any ethical concern. However, as noted in \citet{freq-based-dist}, we believe that distortions in the embedding space may be related to bias in the training data, whether it is inherent to the structure of the modality (e.g. the Zipfian distribution of words), or due to human factors (e.g. geographical considerations).

\section*{Acknowledgements}
This work was funded by the last authors' chair in the PRAIRIE institute funded by the French national agency ANR as part of the ``Investissements d'avenir'' programme under the reference ANR-19-P3IA-0001.

We would like to thank Roman Castagné for useful discussions that led to focusing on observing the effect of anisotropy in the self-attention process.

\bibliography{anthology,custom}

\begin{thebibliography}{38}
\expandafter\ifx\csname natexlab\endcsname\relax\def\natexlab#1{#1}\fi

\bibitem[{Ardila et~al.(2020)Ardila, Branson, Davis, Henretty, Kohler, Meyer,
  Morais, Saunders, Tyers, and Weber}]{commonvoice:2020}
R.~Ardila, M.~Branson, K.~Davis, M.~Henretty, M.~Kohler, J.~Meyer, R.~Morais,
  L.~Saunders, F.~M. Tyers, and G.~Weber. 2020.
\newblock Common voice: A massively-multilingual speech corpus.
\newblock In \emph{Proceedings of the 12th Conference on Language Resources and
  Evaluation (LREC 2020)}, pages 4211--4215.

\bibitem[{Baevski et~al.(2020)Baevski, Zhou, Mohamed, and Auli}]{wav2vec}
Alexei Baevski, Yuhao Zhou, Abdelrahman Mohamed, and Michael Auli. 2020.
\newblock \href
  {https://proceedings.neurips.cc/paper_files/paper/2020/file/92d1e1eb1cd6f9fba3227870bb6d7f07-Paper.pdf}
  {wav2vec 2.0: A framework for self-supervised learning of speech
  representations}.
\newblock In \emph{Advances in Neural Information Processing Systems},
  volume~33, pages 12449--12460. Curran Associates, Inc.

\bibitem[{Bao et~al.(2021)Bao, Dong, and Wei}]{beit-2021}
Hangbo Bao, Li~Dong, and Furu Wei. 2021.
\newblock \href {http://arxiv.org/abs/2106.08254} {Beit: {BERT} pre-training of
  image transformers}.
\newblock \emph{CoRR}, abs/2106.08254.

\bibitem[{Bihani and Rayz(2021)}]{bihani-rayz-2021-low}
Geetanjali Bihani and Julia Rayz. 2021.
\newblock \href {https://doi.org/10.18653/v1/2021.deelio-1.9} {Low anisotropy
  sense retrofitting ({LAS}e{R}) : Towards isotropic and sense enriched
  representations}.
\newblock In \emph{Proceedings of Deep Learning Inside Out (DeeLIO): The 2nd
  Workshop on Knowledge Extraction and Integration for Deep Learning
  Architectures}, pages 81--95, Online. Association for Computational
  Linguistics.

\bibitem[{Bi{\'s} et~al.(2021)Bi{\'s}, Podkorytov, and
  Liu}]{bis-etal-2021-much}
Daniel Bi{\'s}, Maksim Podkorytov, and Xiuwen Liu. 2021.
\newblock \href {https://doi.org/10.18653/v1/2021.naacl-main.403} {Too much in
  common: Shifting of embeddings in transformer language models and its
  implications}.
\newblock In \emph{Proceedings of the 2021 Conference of the North American
  Chapter of the Association for Computational Linguistics: Human Language
  Technologies}, pages 5117--5130, Online. Association for Computational
  Linguistics.

\bibitem[{Clark et~al.(2022)Clark, Garrette, Turc, and
  Wieting}]{clark-etal-2022-canine}
Jonathan~H. Clark, Dan Garrette, Iulia Turc, and John Wieting. 2022.
\newblock \href {https://doi.org/10.1162/tacl_a_00448} {Canine: Pre-training an
  efficient tokenization-free encoder for language representation}.
\newblock \emph{Transactions of the Association for Computational Linguistics},
  10:73--91.

\bibitem[{Clark et~al.(2019)Clark, Khandelwal, Levy, and
  Manning}]{clark-etal-2019-bert}
Kevin Clark, Urvashi Khandelwal, Omer Levy, and Christopher~D. Manning. 2019.
\newblock \href {https://doi.org/10.18653/v1/W19-4828} {What does {BERT} look
  at? an analysis of {BERT}{'}s attention}.
\newblock In \emph{Proceedings of the 2019 ACL Workshop BlackboxNLP: Analyzing
  and Interpreting Neural Networks for NLP}, pages 276--286, Florence, Italy.
  Association for Computational Linguistics.

\bibitem[{Devlin et~al.(2019)Devlin, Chang, Lee, and
  Toutanova}]{devlin-etal-2019-bert}
Jacob Devlin, Ming-Wei Chang, Kenton Lee, and Kristina Toutanova. 2019.
\newblock \href {https://doi.org/10.18653/v1/N19-1423} {{BERT}: Pre-training of
  deep bidirectional transformers for language understanding}.
\newblock In \emph{Proceedings of the 2019 Conference of the North {A}merican
  Chapter of the Association for Computational Linguistics: Human Language
  Technologies, Volume 1 (Long and Short Papers)}, pages 4171--4186,
  Minneapolis, Minnesota. Association for Computational Linguistics.

\bibitem[{El~Boukkouri et~al.(2020)El~Boukkouri, Ferret, Lavergne, Noji,
  Zweigenbaum, and Tsujii}]{el-boukkouri-etal-2020-characterbert}
Hicham El~Boukkouri, Olivier Ferret, Thomas Lavergne, Hiroshi Noji, Pierre
  Zweigenbaum, and Jun{'}ichi Tsujii. 2020.
\newblock \href {https://doi.org/10.18653/v1/2020.coling-main.609}
  {{C}haracter{BERT}: Reconciling {ELM}o and {BERT} for word-level
  open-vocabulary representations from characters}.
\newblock In \emph{Proceedings of the 28th International Conference on
  Computational Linguistics}, pages 6903--6915, Barcelona, Spain (Online).
  International Committee on Computational Linguistics.

\bibitem[{Ethayarajh(2019)}]{ethayarajh-2019-contextual}
Kawin Ethayarajh. 2019.
\newblock \href {https://doi.org/10.18653/v1/D19-1006} {How contextual are
  contextualized word representations? {C}omparing the geometry of {BERT},
  {ELM}o, and {GPT}-2 embeddings}.
\newblock In \emph{Proceedings of the 2019 Conference on Empirical Methods in
  Natural Language Processing and the 9th International Joint Conference on
  Natural Language Processing (EMNLP-IJCNLP)}, pages 55--65, Hong Kong, China.
  Association for Computational Linguistics.

\bibitem[{Gao et~al.(2019)Gao, He, Tan, Qin, Wang, and Liu}]{GaoHTQWL19}
Jun Gao, Di~He, Xu~Tan, Tao Qin, Liwei Wang, and Tie{-}Yan Liu. 2019.
\newblock \href {https://openreview.net/forum?id=SkEYojRqtm} {Representation
  degeneration problem in training natural language generation models}.
\newblock In \emph{7th International Conference on Learning Representations,
  {ICLR} 2019, New Orleans, LA, USA, May 6-9, 2019}. OpenReview.net.

\bibitem[{Gao et~al.(2021)Gao, Yao, and Chen}]{gao-etal-2021-simcse}
Tianyu Gao, Xingcheng Yao, and Danqi Chen. 2021.
\newblock \href {https://doi.org/10.18653/v1/2021.emnlp-main.552} {{S}im{CSE}:
  Simple contrastive learning of sentence embeddings}.
\newblock In \emph{Proceedings of the 2021 Conference on Empirical Methods in
  Natural Language Processing}, pages 6894--6910, Online and Punta Cana,
  Dominican Republic. Association for Computational Linguistics.

\bibitem[{Godey et~al.(2022)Godey, Castagn{\'e}, de~la Clergerie, and
  Sagot}]{godey-etal-2022-manta}
Nathan Godey, Roman Castagn{\'e}, {\'E}ric de~la Clergerie, and Beno{\^\i}t
  Sagot. 2022.
\newblock \href {https://aclanthology.org/2022.findings-emnlp.207} {{MANT}a:
  Efficient gradient-based tokenization for end-to-end robust language
  modeling}.
\newblock In \emph{Findings of the Association for Computational Linguistics:
  EMNLP 2022}, pages 2859--2870, Abu Dhabi, United Arab Emirates. Association
  for Computational Linguistics.

\bibitem[{Guo et~al.(2022)Guo, Lu, Liu, Cheng, and Hu}]{guo2022visual}
Meng-Hao Guo, Cheng-Ze Lu, Zheng-Ning Liu, Ming-Ming Cheng, and Shi-Min Hu.
  2022.
\newblock Visual attention network.
\newblock \emph{arXiv preprint arXiv:2202.09741}.

\bibitem[{He et~al.(2016)He, Zhang, Ren, and Sun}]{he2016deep}
Kaiming He, Xiangyu Zhang, Shaoqing Ren, and Jian Sun. 2016.
\newblock Deep residual learning for image recognition.
\newblock In \emph{Proceedings of the IEEE conference on computer vision and
  pattern recognition}, pages 770--778.

\bibitem[{Hsu et~al.(2021)Hsu, Bolte, Tsai, Lakhotia, Salakhutdinov, and
  Mohamed}]{HuBERT}
Wei-Ning Hsu, Benjamin Bolte, Yao-Hung~Hubert Tsai, Kushal Lakhotia, Ruslan
  Salakhutdinov, and Abdelrahman Mohamed. 2021.
\newblock \href {https://doi.org/10.1109/TASLP.2021.3122291} {Hubert:
  Self-supervised speech representation learning by masked prediction of hidden
  units}.
\newblock \emph{IEEE/ACM Transactions on Audio, Speech, and Language
  Processing}, 29:3451--3460.

\bibitem[{Ioffe and Szegedy(2015)}]{pmlr-v37-ioffe15}
Sergey Ioffe and Christian Szegedy. 2015.
\newblock \href {https://proceedings.mlr.press/v37/ioffe15.html} {Batch
  normalization: Accelerating deep network training by reducing internal
  covariate shift}.
\newblock In \emph{Proceedings of the 32nd International Conference on Machine
  Learning}, volume~37 of \emph{Proceedings of Machine Learning Research},
  pages 448--456, Lille, France. PMLR.

\bibitem[{Liu et~al.(2019)Liu, Ott, Goyal, Du, Joshi, Chen, Levy, Lewis,
  Zettlemoyer, and Stoyanov}]{roberta}
Yinhan Liu, Myle Ott, Naman Goyal, Jingfei Du, Mandar Joshi, Danqi Chen, Omer
  Levy, Mike Lewis, Luke Zettlemoyer, and Veselin Stoyanov. 2019.
\newblock \href {http://arxiv.org/abs/1907.11692} {Roberta: {A} robustly
  optimized {BERT} pretraining approach}.
\newblock \emph{CoRR}, abs/1907.11692.

\bibitem[{Liu et~al.(2022)Liu, Mao, Wu, Feichtenhofer, Darrell, and
  Xie}]{liu2022convnet}
Zhuang Liu, Hanzi Mao, Chao-Yuan Wu, Christoph Feichtenhofer, Trevor Darrell,
  and Saining Xie. 2022.
\newblock A convnet for the 2020s.
\newblock \emph{Proceedings of the IEEE/CVF Conference on Computer Vision and
  Pattern Recognition (CVPR)}.

\bibitem[{Merity et~al.(2016)Merity, Xiong, Bradbury, and Socher}]{MerityXBS16}
Stephen Merity, Caiming Xiong, James Bradbury, and Richard Socher. 2016.
\newblock \href {http://arxiv.org/abs/1609.07843} {Pointer sentinel mixture
  models}.
\newblock \emph{CoRR}, abs/1609.07843.

\bibitem[{Puccetti et~al.(2022)Puccetti, Rogers, Drozd, and
  Dell{'}Orletta}]{puccetti-etal-2022-outlier}
Giovanni Puccetti, Anna Rogers, Aleksandr Drozd, and Felice Dell{'}Orletta.
  2022.
\newblock \href {https://aclanthology.org/2022.findings-emnlp.93} {Outlier
  dimensions that disrupt transformers are driven by frequency}.
\newblock In \emph{Findings of the Association for Computational Linguistics:
  EMNLP 2022}, pages 1286--1304, Abu Dhabi, United Arab Emirates. Association
  for Computational Linguistics.

\bibitem[{Radford et~al.(2022)Radford, Kim, Xu, Brockman, McLeavey, and
  Sutskever}]{radford2022whisper}
Alec Radford, Jong~Wook Kim, Tao Xu, Greg Brockman, Christine McLeavey, and
  Ilya Sutskever. 2022.
\newblock \href {https://doi.org/10.48550/ARXIV.2212.04356} {Robust speech
  recognition via large-scale weak supervision}.

\bibitem[{Radford et~al.(2019)Radford, Wu, Child, Luan, Amodei, and
  Sutskever}]{gpt2}
Alec Radford, Jeff Wu, Rewon Child, David Luan, Dario Amodei, and Ilya
  Sutskever. 2019.
\newblock Language models are unsupervised multitask learners.

\bibitem[{Raffel et~al.(2020)Raffel, Shazeer, Roberts, Lee, Narang, Matena,
  Zhou, Li, and Liu}]{2020t5}
Colin Raffel, Noam Shazeer, Adam Roberts, Katherine Lee, Sharan Narang, Michael
  Matena, Yanqi Zhou, Wei Li, and Peter~J. Liu. 2020.
\newblock \href {http://jmlr.org/papers/v21/20-074.html} {Exploring the limits
  of transfer learning with a unified text-to-text transformer}.
\newblock \emph{Journal of Machine Learning Research}, 21(140):1--67.

\bibitem[{Rajaee and Pilehvar(2021)}]{rajaee-pilehvar-2021-cluster}
Sara Rajaee and Mohammad~Taher Pilehvar. 2021.
\newblock \href {https://doi.org/10.18653/v1/2021.acl-short.73} {A
  cluster-based approach for improving isotropy in contextual embedding space}.
\newblock In \emph{Proceedings of the 59th Annual Meeting of the Association
  for Computational Linguistics and the 11th International Joint Conference on
  Natural Language Processing (Volume 2: Short Papers)}, pages 575--584,
  Online. Association for Computational Linguistics.

\bibitem[{Rajaee and Pilehvar(2022)}]{rajaee-pilehvar-2022-isotropy}
Sara Rajaee and Mohammad~Taher Pilehvar. 2022.
\newblock \href {https://doi.org/10.18653/v1/2022.findings-acl.103} {An
  isotropy analysis in the multilingual {BERT} embedding space}.
\newblock In \emph{Findings of the Association for Computational Linguistics:
  ACL 2022}, pages 1309--1316, Dublin, Ireland. Association for Computational
  Linguistics.

\bibitem[{Russakovsky et~al.(2015)Russakovsky, Deng, Su, Krause, Satheesh, Ma,
  Huang, Karpathy, Khosla, Bernstein, Berg, and
  Fei-Fei}]{imagenet15russakovsky}
Olga Russakovsky, Jia Deng, Hao Su, Jonathan Krause, Sanjeev Satheesh, Sean Ma,
  Zhiheng Huang, Andrej Karpathy, Aditya Khosla, Michael Bernstein,
  Alexander~C. Berg, and Li~Fei-Fei. 2015.
\newblock \href {https://doi.org/10.1007/s11263-015-0816-y} {{ImageNet Large
  Scale Visual Recognition Challenge}}.
\newblock \emph{International Journal of Computer Vision (IJCV)},
  115(3):211--252.

\bibitem[{Su et~al.(2021)Su, Cao, Liu, and Ou}]{su2021whitening}
Jianlin Su, Jiarun Cao, Weijie Liu, and Yangyiwen Ou. 2021.
\newblock Whitening sentence representations for better semantics and faster
  retrieval.
\newblock \emph{arXiv preprint arXiv:2103.15316}.

\bibitem[{Tan and Le(2019)}]{Tan2019EfficientNetRM}
Mingxing Tan and Quoc~V. Le. 2019.
\newblock Efficientnet: Rethinking model scaling for convolutional neural
  networks.
\newblock \emph{ArXiv}, abs/1905.11946.

\bibitem[{Touvron et~al.(2021)Touvron, Cord, Douze, Massa, Sablayrolles, and
  Jegou}]{pmlr-v139-touvron21a}
Hugo Touvron, Matthieu Cord, Matthijs Douze, Francisco Massa, Alexandre
  Sablayrolles, and Herve Jegou. 2021.
\newblock \href {https://proceedings.mlr.press/v139/touvron21a.html} {Training
  data-efficient image transformers and distillation through attention}.
\newblock In \emph{Proceedings of the 38th International Conference on Machine
  Learning}, volume 139 of \emph{Proceedings of Machine Learning Research},
  pages 10347--10357. PMLR.

\bibitem[{Wang et~al.(2020)Wang, Huang, Huang, Hu, Wang, and
  Gu}]{Wang2020Improving}
Lingxiao Wang, Jing Huang, Kevin Huang, Ziniu Hu, Guangtao Wang, and Quanquan
  Gu. 2020.
\newblock \href {https://openreview.net/forum?id=ByxY8CNtvr} {Improving neural
  language generation with spectrum control}.
\newblock In \emph{International Conference on Learning Representations}.

\bibitem[{Wu et~al.(2020)Wu, Xu, Dai, Wan, Zhang, Tomizuka, Keutzer, and
  Vajda}]{Wu2020VisualTT}
Bichen Wu, Chenfeng Xu, Xiaoliang Dai, Alvin Wan, Peizhao Zhang, Masayoshi
  Tomizuka, Kurt Keutzer, and P{\'e}ter Vajda. 2020.
\newblock Visual transformers: Token-based image representation and processing
  for computer vision.
\newblock \emph{ArXiv}, abs/2006.03677.

\bibitem[{Wu et~al.(2021)Wu, Xiao, Codella, Liu, Dai, Yuan, and
  Zhang}]{wu2021cvt}
Haiping Wu, Bin Xiao, Noel Codella, Mengchen Liu, Xiyang Dai, Lu~Yuan, and Lei
  Zhang. 2021.
\newblock Cvt: Introducing convolutions to vision transformers.
\newblock \emph{arXiv preprint arXiv:2103.15808}.

\bibitem[{Xie et~al.(2021)Xie, Wang, Yu, Anandkumar, Alvarez, and
  Luo}]{segformer21}
Enze Xie, Wenhai Wang, Zhiding Yu, Anima Anandkumar, Jose~M. Alvarez, and Ping
  Luo. 2021.
\newblock \href {http://arxiv.org/abs/2105.15203} {Segformer: Simple and
  efficient design for semantic segmentation with transformers}.
\newblock \emph{CoRR}, abs/2105.15203.

\bibitem[{Xue et~al.(2022)Xue, Barua, Constant, Al-Rfou, Narang, Kale, Roberts,
  and Raffel}]{xue-etal-2022-byt5}
Linting Xue, Aditya Barua, Noah Constant, Rami Al-Rfou, Sharan Narang, Mihir
  Kale, Adam Roberts, and Colin Raffel. 2022.
\newblock \href {https://doi.org/10.1162/tacl_a_00461} {{B}y{T}5: Towards a
  token-free future with pre-trained byte-to-byte models}.
\newblock \emph{Transactions of the Association for Computational Linguistics},
  10:291--306.

\bibitem[{Yan et~al.(2021)Yan, Li, Wang, Zhang, Wu, and
  Xu}]{yan-etal-2021-consert}
Yuanmeng Yan, Rumei Li, Sirui Wang, Fuzheng Zhang, Wei Wu, and Weiran Xu. 2021.
\newblock \href {https://doi.org/10.18653/v1/2021.acl-long.393} {{C}on{SERT}: A
  contrastive framework for self-supervised sentence representation transfer}.
\newblock In \emph{Proceedings of the 59th Annual Meeting of the Association
  for Computational Linguistics and the 11th International Joint Conference on
  Natural Language Processing (Volume 1: Long Papers)}, pages 5065--5075,
  Online. Association for Computational Linguistics.

\bibitem[{Yu et~al.(2022)Yu, Song, Kim, Lee, Ryu, and Yoon}]{yu-etal-2022-rare}
Sangwon Yu, Jongyoon Song, Heeseung Kim, Seongmin Lee, Woo-Jong Ryu, and
  Sungroh Yoon. 2022.
\newblock \href {https://doi.org/10.18653/v1/2022.acl-long.3} {Rare tokens
  degenerate all tokens: Improving neural text generation via adaptive gradient
  gating for rare token embeddings}.
\newblock In \emph{Proceedings of the 60th Annual Meeting of the Association
  for Computational Linguistics (Volume 1: Long Papers)}, pages 29--45, Dublin,
  Ireland. Association for Computational Linguistics.

\bibitem[{Zhou et~al.(2021)Zhou, Ethayarajh, and Jurafsky}]{freq-based-dist}
Kaitlyn Zhou, Kawin Ethayarajh, and Dan Jurafsky. 2021.
\newblock \href {http://arxiv.org/abs/2104.08465} {Frequency-based distortions
  in contextualized word embeddings}.
\newblock \emph{CoRR}, abs/2104.08465.

\end{thebibliography}
\bibliographystyle{acl_natbib}

\appendix

\section{Pearson correlation of the drift norm and anisotropy}
\label{sec:pearson}

\begin{figure}[H]
    \centering
    \includegraphics[width=\linewidth]{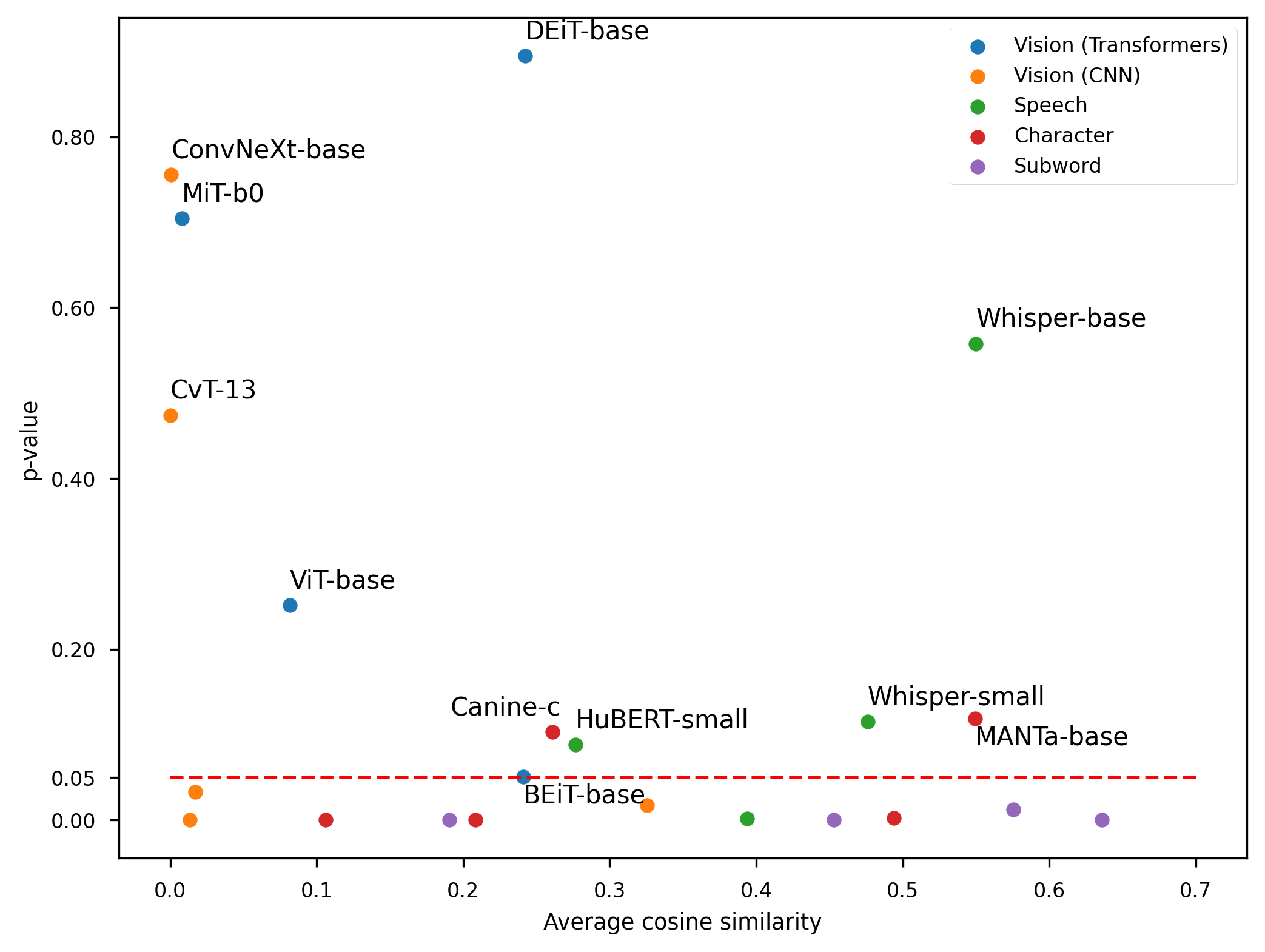}
    \caption{p-value of the Pearson correlation test between the norm of the average representation and the cosine-similarity averaged over all layers, across modalities. Models above the red dotted line are not significantly affected by the drift effect.}
    \label{fig:pval_vs_cos_pearson}
\end{figure}

The Pearson test measures a linear correlation between random variables, while the Spearman test measures a monotonic correlation. As there is no specific argument in favor of a linear relationship between the measured distributions (average cosine-similarity and norm of the average representation), we decided to use the Spearman correlation test in order to take into account more complex relation patterns.

Nevertheless, this metric is based on the rank of each observation, and is thus not robust to fluctuations due to sample variance, specifically when working with such small samples. This is reflected by the discrepancy between Pearson and Spearman p-values for some models (e.g. GPT-2).

\end{document}